\documentclass[11pt]{article}

\usepackage[preprint]{acl}

\usepackage{times}
\usepackage{latexsym}
\usepackage[T1]{fontenc}
\usepackage[utf8]{inputenc}
\usepackage{microtype}
\usepackage{inconsolata}

\usepackage{url}
\usepackage{booktabs}
\usepackage{colortbl}
\usepackage{amsmath,amssymb,amsthm}
\usepackage{xspace}
\usepackage{enumitem}

\usepackage{graphicx}
\usepackage[skins,breakable,listings]{tcolorbox}

\definecolor{tablegroupblue}{RGB}{235,250,252}
\definecolor{rankfirstblue}{RGB}{220,235,248}
\definecolor{ranksecondgreen}{RGB}{230,246,234}
\definecolor{rankthirdpeach}{RGB}{251,240,224}
\definecolor{prompttitleblue}{RGB}{69,108,137}
\definecolor{promptbodyblue}{RGB}{245,250,253}

\newtcblisting{promptbox}[1]{
  enhanced,
  breakable,
  colback=promptbodyblue,
  colframe=prompttitleblue,
  colbacktitle=prompttitleblue,
  coltitle=white,
  title={#1},
  fonttitle=\bfseries\raggedright,
  width=\columnwidth,
  listing only,
  listing options={
    basicstyle=\ttfamily\scriptsize,
    breaklines=true,
    breakatwhitespace=true,
    breakautoindent=false,
    breakindent=0pt,
    columns=fullflexible,
    keepspaces=true,
    showstringspaces=false
  },
  boxrule=0.8pt,
  arc=5pt,
  left=5pt,
  right=5pt,
  top=5pt,
  bottom=5pt,
  toptitle=3pt,
  bottomtitle=3pt,
  before skip=6pt,
  after skip=8pt
}

\setlist[itemize]{leftmargin=*, nosep}
\setlist[enumerate]{leftmargin=*, nosep}

\newcommand{\dataset}{{\fontfamily{lmtt}\selectfont\textbf{MedLoCoMo}}\xspace}

\title{MedLoCoMo: A Long-Context Multi-Session Medical Dialogue Benchmark for Large Language Models}

\author{
  Zeyu Zhang \quad Ziqing Wang \quad Kaize Ding \\
  Northwestern University \\
  \texttt{\{zeyuzhang2030, ziqingwang2029\}@u.northwestern.edu} \\
  \texttt{kaize.ding@northwestern.edu}
}

\begin{document}

\maketitle

\begin{abstract}
\dataset is a \textbf{Med}ical \textbf{Lo}ng-\textbf{Co}ntext Me\textbf{mo}ry benchmark for patient-specific clinical reasoning over multi-admission medical dialogue. Existing medical QA benchmarks largely test short-context knowledge or single-document grounding, leaving open whether LLMs can use, connect, and abstain over longitudinal patient histories. We build \dataset from deidentified MIMIC-IV and MIMIC-IV-Note records by constructing admission-level clinical packets, synthesizing grounded doctor--patient conversations, and generating evidence-linked QA items over single-admission, cross-admission, and adversarial unanswerable settings. The benchmark contains 100 patient timelines averaging 1,669.8 turns, 29.7 sessions, and 74,512.2 tokens per conversation. Across the evaluated baselines, cross-admission reasoning is consistently harder than localized evidence use, even when models have long context windows or use external memory or retrieval methods. The code and \dataset benchmark release is available at \url{https://github.com/leozzy13/MedLoCoMo} for use and reproducibility.
\end{abstract}

\begin{figure}[t]
  \centering
  \includegraphics[width=\columnwidth]{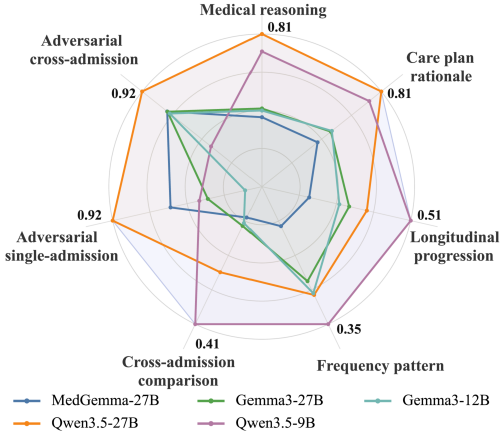}
  \caption{\textbf{Radar plot of selected model performance across question types in \dataset.} Axes report LLM-judge accuracy for answerable question types and abstention accuracy for adversarial types. Scores are shown on a 0--1 scale.}
  \label{fig:radar_question_type_scores}
\end{figure}

\begin{table*}[t]
  \centering
  \small
  \setlength{\tabcolsep}{4pt}
  \caption{\textbf{Comparison of \dataset with long-context, dialogue, and clinical benchmark datasets.} Columns summarize medical domain coverage, dataset scale, average conversation length, session structure, token count, and temporal span where those fields are defined; \dataset combines medical grounding with long multi-session patient timelines for evaluation.}
  \label{tab:dataset_comparison}
  \resizebox{\textwidth}{!}{%
  \begin{tabular}{lcccccc}
    \toprule
    Dataset & Medical & \# Conv./cases & Avg. turns/conv. & Avg. sessions/conv. & Avg. tokens/conv. & Time interval \\
    \midrule
    LoCoMo~\citep{maharana-etal-2024-evaluating} & No & 10 & 588.2 & 27.2 & 16,618.1 & few months \\
    MSC~\citep{xu-etal-2022-beyond} & No & 1,001 & 53.3 & 4 & 1,225.9 & few days \\
    MPChat~\citep{ahn-etal-2023-mpchat} & No & 15,000 & 2.8 & 1 & 53.3 & -- \\
    MedDialog-EN~\citep{zeng-etal-2020-meddialog} & Yes & 257,332 & 2 & 1 & 173.0 & -- \\
    ACI-Bench~\citep{yim2023acibench} & Yes & 207 & 55 & 1 & 1,302 & few minutes \\
    \midrule
    \textbf{\dataset (ours)} & \textbf{Yes} & \textbf{100} & \textbf{1,669.8} & \textbf{29.7} & \textbf{74,512.2} & \textbf{few years} \\
    \bottomrule
  \end{tabular}%
  }
\end{table*}


\section{Introduction}
\label{sec:introduction}

Clinical reasoning is inherently longitudinal. Patient-specific questions often require evidence from prior admissions, discharge summaries, imaging reports, procedures, microbiology results, and evolving problem lists. These records form a heterogeneous sequence of clinical events rather than a single passage, making deidentified EHR resources such as MIMIC-IV~\citep{johnson2023mimiciv} important for clinical NLP research.

The central evaluation gap is patient-level longitudinal reasoning. Exam-style and biomedical QA datasets such as MedQA, MedMCQA, and PubMedQA measure domain knowledge from compact questions, vignettes, or abstracts~\citep{jin2020disease,pmlr-v174-pal22a,jin-etal-2019-pubmedqa}. EHR-grounded QA datasets such as emrQA and EHRNoteQA move closer to clinical documentation, but still mainly focus on single notes or bounded note contexts~\citep{pampari-etal-2018-emrqa,kweon2024ehrnoteqa}. They do not directly test whether LLMs can answer patient-specific questions whose evidence spans many admissions, or abstain when the longitudinal record lacks support.

This gap is challenging because longitudinal clinical QA stresses more than context length. Relevant evidence can be sparse, repeated, or updated across admissions, requiring models to decide which earlier events still matter and which later events change the patient's state. Prior long-context studies show degradation as contexts grow or relevant spans move away from salient prompt positions~\citep{bai-etal-2024-longbench,liu-etal-2024-lost}. In patient timelines, these failures appear as missed cross-admission links, stale evidence use, or unsupported answers to plausible questions. A useful benchmark should therefore test both evidence use across encounters and abstention when no admission supports the requested fact~\citep{rajpurkar-etal-2018-know}. Figure~\ref{fig:radar_question_type_scores} previews this variation across \dataset question types.

We introduce \dataset, a \textbf{Med}ical \textbf{Lo}ng-\textbf{Co}ntext Me\textbf{mo}ry benchmark for patient-specific clinical reasoning. Starting from deidentified MIMIC-IV structured data and MIMIC-IV-Note text~\citep{johnson2023mimiciv,johnson2023mimicivnote}, we construct admission-level clinical packets, synthesize grounded doctor--patient style conversations, and generate short-answer QA items with explicit evidence references. Each benchmark instance is a chronological patient timeline whose admissions are treated as sessions in one long conversation. \dataset is fixed before evaluation, so memory and retrieval systems are compared afterward on the same evidence-linked QA set. The resulting benchmark contains 100 patient conversations, averaging 1,669.8 turns, 29.7 sessions, and 74,512.2 tokens per conversation.

Our evaluation separates three sources of difficulty: \emph{single-admission} questions answerable from one hospitalization, \emph{cross-admission} questions requiring evidence from multiple admissions, and \emph{adversarial unanswerable} questions that test abstention when plausible clinical questions lack support. Because long patient timelines also stress evidence preservation and updating, we include a memory-augmentation case study with retrieval methods as additional baselines. Figure~\ref{fig:pipeline_overview} summarizes the pipeline; Appendices~\ref{app:medlocomo_details}--\ref{app:generation_prompts} provide additional dataset, evaluation, and prompt details.

Our contributions are:
\begin{itemize}
    \item \textbf{A patient-level long-context clinical benchmark} built from deidentified EHR data and notes, with 100 longitudinal patient timelines averaging 1,669.8 turns and 74,512.2 tokens.
    \item \textbf{An evidence-linked QA design} that separates single-admission reasoning, cross-admission reasoning, and adversarial unanswerable questions.
    \item \textbf{A baseline study} of general-purpose and medical LLMs showing that, in the evaluated suite, model scale and domain specialization do not remove the cross-admission reasoning gap.
    \item \textbf{A memory-augmented case study} with retrieval baselines, showing that memory can improve abstention and aggregate QA Score but richer memory structure alone does not guarantee better answer extraction in this setting.
\end{itemize}

\section{Related Work}
\label{sec:related_work}

\paragraph{Medical QA Benchmarks.}
Medical QA benchmarks largely measure domain knowledge in compact settings: MedQA~\citep{jin2020disease} and MedMCQA~\citep{pmlr-v174-pal22a} use exam-style questions, while PubMedQA~\citep{jin-etal-2019-pubmedqa} and BioASQ~\citep{tsatsaronis2015bioasq} evaluate biomedical literature understanding. EHR-grounded QA moves closer to clinical documentation: emrQA constructs question-answer pairs from clinical notes~\citep{pampari-etal-2018-emrqa}, and EHRNoteQA evaluates questions grounded in discharge summaries~\citep{kweon2024ehrnoteqa}. MedOdyssey targets medical long-context evaluation up to 200K tokens~\citep{fan2024medodyssey}, but is organized around long medical inputs and retrieval-style stress tests rather than patient-level, multi-admission dialogue. In contrast, \dataset treats the patient as the unit of evaluation and tests evidence use across a longitudinal timeline.

\paragraph{Long-Context QA Evaluation.}
Long-context QA evaluation asks whether models can locate, retain, and combine evidence across thousands of tokens. LongBench~\citep{bai-etal-2024-longbench}, LongBench v2~\citep{bai-etal-2025-longbench}, and L-Eval~\citep{an-etal-2024-l} include long-document QA, summarization, dialogue understanding, and deeper reasoning tasks, while RULER probes controlled reasoning at different context lengths~\citep{hsieh2024ruler}. Positional-robustness work shows that models can fail when evidence appears in less salient context positions~\citep{liu-etal-2024-lost}, so longer prompts alone are not enough. Long patient histories also resemble long-term conversational memory problems: LoCoMo evaluates very long multi-session conversations~\citep{maharana-etal-2024-evaluating}; MedMT-Bench tests memory, interference robustness, and safety defense in synthetic long multi-turn medical scenarios~\citep{yang2026medmt}; MSC studies repeated interactions over time~\citep{xu-etal-2022-beyond}; and recent analysis suggests that multi-turn assumptions and errors can compound~\citep{laban2026llms}. \dataset follows this multi-session framing, but grounds each session in patient-specific EHR evidence.

\paragraph{Clinical Dialogue Generation.}
Medical dialogue corpora such as MedDialog~\citep{zeng-etal-2020-meddialog} and \(M^2\)-MedDialog~\citep{yan2021m2meddialog} support healthcare conversation modeling, but online consultations are not typically linked to structured EHR timelines. ACI-Bench focuses on visit-note generation from encounter dialogue~\citep{yim2023acibench}, while MedDialogRubrics evaluates multi-turn medical consultations with rubric-based assessment~\citep{gong2026meddialogrubrics}. MediLongChat synthesizes knowledge-guided, multi-encounter medical dialogues for in-dialogue, cross-dialogue, and synthesis reasoning~\citep{hu2026medilongchat}; \dataset instead starts from deidentified EHR-derived admissions and emphasizes evidence-linked QA over long patient timelines. Prior work such as Self-Instruct shows that LLM-based generation can scale instruction-style data construction~\citep{wang-etal-2023-self-instruct}; in clinical benchmarking, this makes grounding, evidence references, and unsupported-answer cases essential.

\paragraph{How MedLoCoMo Differs.}
Table~\ref{tab:dataset_comparison} positions \dataset against long-context, dialogue, and clinical benchmarks by medical relevance, scale, session structure, and context length. Across 100 conversations, \dataset averages 1,669.8 turns, 29.7 sessions, and 74,512.2 tokens; the longest conversation reaches 156,528 tokens and 64 sessions. Compared with short-context medical QA, single-note EHR QA, medical long-context document benchmarks, and consultation benchmarks, \dataset emphasizes evidence use across admissions, evolving clinical facts, and abstention when the record does not support an answer~\citep{rajpurkar-etal-2018-know}. Retrieval-augmented generation~\citep{lewis2020retrieval}, medical RAG benchmarks such as MIRAGE~\citep{xiong-etal-2024-benchmarking}, and clinical temporal information extraction benchmarks such as i2b2~\citep{sun2013evaluating} offer complementary baselines for evidence selection and event structuring, but they do not by themselves evaluate open-answer dialogue QA over complete patient timelines.

\section{\dataset Generation Pipeline}
\label{sec:dataset_generation}
\begin{figure*}[t]
  \centering
  \includegraphics[width=\textwidth]{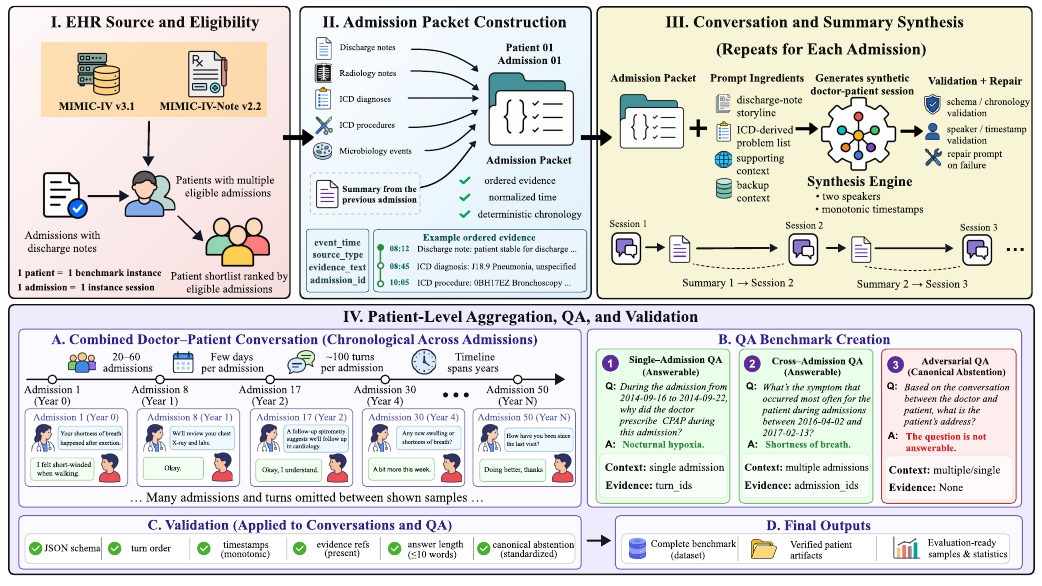}
  \caption{\textbf{End-to-end construction pipeline for \dataset.} Eligible MIMIC-IV and MIMIC-IV-Note records are converted into admission packets, used to synthesize grounded doctor--patient conversations and admission summaries, assembled into patient-level timelines, and converted into evidence-linked QA items for downstream evaluation and analysis.}
  \label{fig:pipeline_overview}
\end{figure*}

\dataset is constructed through a staged pipeline that converts source EHR records into patient-level long-context QA instances. We first identify eligible longitudinal patient trajectories, then form admission-level clinical packets from notes and structured events. These packets support grounded conversation and summary synthesis, chronological patient-level aggregation, evidence-linked QA generation, and final quality-control checks. Figure~\ref{fig:pipeline_overview} summarizes this process.

\subsection{Source Data and Eligibility}
\dataset is built from MIMIC-IV v3.1 and MIMIC-IV-Note v2.2, which provide deidentified structured EHR data and linked free-text clinical notes~\citep{johnson2023mimiciv,johnson2023mimicivnote}. The benchmark is organized at the patient level. Each patient is one benchmark instance, and each hospital admission is one session in that instance. Concatenating the sessions chronologically creates a multi-session, multi-turn long-context input.

\paragraph{Eligible admissions.}
An admission is eligible if it contains at least one discharge summary in MIMIC-IV-Note. We require discharge summaries because they provide the most complete narrative of the hospital course and serve as the backbone for grounded dialogue generation. Admissions without discharge notes are excluded.

\paragraph{Eligible patients.}
A patient is eligible if they have multiple eligible admissions. To support large-scale generation, we use a curated shortlist that ranks patients by the number of eligible admissions. This shortlist favors longitudinal trajectories with many encounters, which are especially useful for stress-testing long-context reasoning.

\subsection{Clinical Packet Formation}
For each eligible admission, we construct an \emph{admission packet}: a structured JSON artifact that gathers the clinical evidence used for generation. Each packet contains:
\begin{itemize}
    \setlength{\itemsep}{2pt}
    \setlength{\parsep}{0pt}
    \setlength{\topsep}{2pt}
    \item \textbf{Backbone narrative:} discharge notes for the admission, ordered by note sequence and time.
    \item \textbf{Supporting free text:} radiology notes linked to the same admission.
    \item \textbf{Structured clinical events:} admission-level ICD diagnoses, procedures, and microbiology entries.
    \item \textbf{Longitudinal context:} the generated summary from the previous admission, when available, to support cross-session continuity.
\end{itemize}

Timestamp handling for date-only fields and admission boundaries is described in Appendix~\ref{app:timestamp_normalization} for reproducible evaluation.

\subsection{Conversation and Summary Synthesis}
Given an admission packet, we prompt an LLM to generate a grounded doctor--patient style conversation for that admission. We enforce the following constraints through prompts and post-generation validation:
\begin{itemize}
    \setlength{\itemsep}{2pt}
    \setlength{\parsep}{0pt}
    \setlength{\topsep}{2pt}
    \item Exactly two speakers: \textbf{Doctor} and \textbf{Patient}.
    \item The output is a list of timestamped turns with contiguous turn indices.
    \item Turn timestamps are monotonically increasing and fall within the admission boundaries.
    \item The discharge notes provide the backbone storyline; radiology, procedures, microbiology, and previous-admission summaries will provide the supporting context.
\end{itemize}

\begin{figure*}[!t]
  \centering
  \begin{minipage}{0.3492\textwidth}
    \centering
    \includegraphics[width=\linewidth]{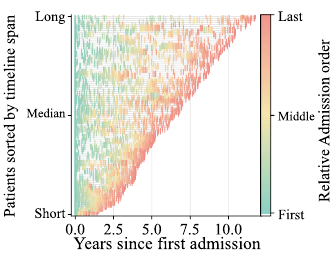}\par\vspace{2pt}
    {\small\textbf{(a) Longitudinal Coverage}}
  \end{minipage}
  \hfill
  \begin{minipage}{0.3492\textwidth}
    \centering
    \includegraphics[width=\linewidth]{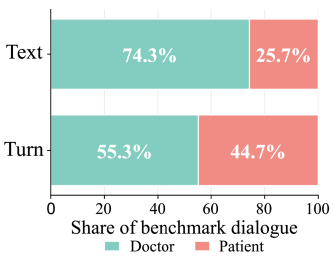}\par\vspace{2pt}
    {\small\textbf{(b) Dialogue Composition}}
  \end{minipage}
  \hfill
  \begin{minipage}{0.28162\textwidth}
    \centering
    \includegraphics[width=\linewidth]{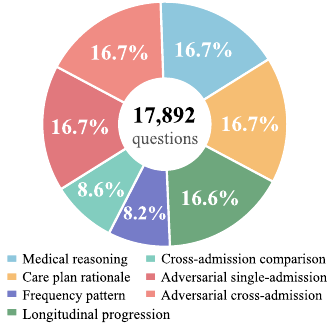}\par\vspace{2pt}
    {\small\textbf{(c) Question-Type Distribution}}
  \end{minipage}
  \caption{\textbf{Benchmark overview for \dataset.} Panel (a) visualizes longitudinal coverage across patient admissions, panel (b) reports the doctor--patient composition of generated dialogue by text and turn share, and panel (c) shows the QA type mix spanning single-admission, cross-admission, and adversarial unanswerable questions.}
  \label{fig:benchmark_overview}
\end{figure*}

\paragraph{Per-admission summary.}
Alongside conversation lines, the model generates a structured admission summary containing admission boundaries, a hospital-course paragraph, and a problem list. The prompt requires the summary to be derived only from the generated conversation rather than directly from the packet, reducing direct reuse of note phrasing in downstream summary-based evaluation.

\paragraph{Chronological multi-session generation.}
Admissions are generated in chronological order. After each admission, its generated summary is passed to the next admission as lightweight longitudinal context. This mechanism encourages cross-session consistency, such as preserving chronic conditions or recurring problems, while avoiding repeated ingestion of the full prior dialogue.

\subsection{Patient-Level Aggregation}
After generating all eligible admissions for a patient, we assemble patient-level artifacts that define the long-context benchmark input. Appendix~\ref{app:medlocomo_details} describes the released artifact layout. The main patient-level artifacts are:
\begin{itemize}
    \setlength{\itemsep}{2pt}
    \setlength{\parsep}{0pt}
    \setlength{\topsep}{2pt}
    \item \textbf{Combined conversation}: a chronological list of admission sessions, each containing its generated conversation lines.
    \item \textbf{Patient summary}: bookkeeping metadata and generation statistics, including processed versus eligible admissions, for reproducibility.
\end{itemize}

\subsection{QA Generation}
After patient-level timelines are assembled, the final generation stage converts them into evidence-linked short-answer QA items. We generate questions at two evidence scopes: \emph{single-admission}, where the answer is supported by one session, and \emph{cross-admission}, where the answer requires multiple sessions. Each QA item includes id, scope, question type, question, answer, and evidence. Answers are constrained to at most 10 words. Figure~\ref{fig:benchmark_overview} summarizes the benchmark timeline, dialogue composition, and resulting distribution across clinical question types in the benchmark.

\paragraph{Single-admission QA.}
For each admission conversation, we generate three QA pairs: two answerable questions and one adversarial unanswerable question. Answerable questions are grounded in the admission conversation and cover medical reasoning and care plan rationale. The unanswerable question uses the adversarial type and stores the standard gold answer ``the question is not answerable.'' Evidence cites the admission identifier and supporting turn identifiers.

\paragraph{Cross-admission QA.}
Cross-admission questions target patient-level reasoning across sessions. Answerable questions cover longitudinal progression, cross-admission comparison, and frequency pattern, and must cite evidence from at least two admissions. Adversarial cross-admission questions use the same standard abstention gold answer, but they draw on multiple admissions to create plausible unsupported longitudinal questions.

\paragraph{Human quality review.}
As the full benchmark is too large for complete manual inspection, we conducted a sampled human audit of about 50 generated admission conversations against their source EHR packets and about 300 QA items across single-admission, cross-admission, and adversarial settings. The review checked factual faithfulness of conversation content to the EHR-derived evidence, correctness and sufficiency of cited evidence, answerability for answerable QA, and true unanswerability for adversarial QA. All human-reviewed cases passed these checks. For the full benchmark, we apply automated quality-control checks before release; Appendix~\ref{app:medlocomo_details} provides details.

\begin{table*}[!t]
  \centering
  \small
  \setlength{\tabcolsep}{3pt}
  \caption{\textbf{Baseline model results on \dataset.} F1 and LLM-judge accuracy (J) evaluate answerable questions, and adversarial accuracy (Acc) evaluates unanswerable questions. Score is the item-weighted combination of J on answerable items and Acc on adversarial items within the corresponding split; all values are percentages. In Score columns, \protect\colorbox{rankfirstblue}{\protect\textbf{first}}, \protect\colorbox{ranksecondgreen}{\protect\underline{second}}, and \protect\colorbox{rankthirdpeach}{third} mark summary ranks within each model group.}
  \label{tab:baseline_model_results}
  \resizebox{\textwidth}{!}{%
  \begin{tabular}{@{}l@{\hspace{4pt}}c@{\hspace{18pt}}cccc@{\hspace{18pt}}cccc@{\hspace{18pt}}cccc@{}}
    \toprule
    \multicolumn{1}{@{}l}{\smash{\raisebox{-0.7\normalbaselineskip}{\textbf{Model}}}} & \multicolumn{1}{@{\hspace{1pt}}c@{\hspace{8pt}}}{\smash{\raisebox{-1.15\normalbaselineskip}{\shortstack[c]{\textbf{Context}\\\textbf{Length}}}}} & \multicolumn{4}{c}{\textbf{Overall}} & \multicolumn{4}{c}{\textbf{Single-admission}} & \multicolumn{4}{c}{\textbf{Cross-admission}} \\
    \cmidrule(lr){3-6}\cmidrule(lr){7-10}\cmidrule(l){11-14}
    & & F1 & J & Acc & Score & F1 & J & Acc & Score & F1 & J & Acc & Score \\
    \midrule
    \rowcolor{tablegroupblue}\multicolumn{14}{c}{\rule[-0.75ex]{0pt}{2.4ex}\textit{General-Purpose LLMs}} \\
    GPT-5.1 & 400K & 21.1 & 64.2 & 90.2 & \cellcolor{rankfirstblue}\textbf{72.8} & 28.2 & 88.4 & 89.0 & \cellcolor{rankfirstblue}\textbf{88.6} & 14.1 & 39.9 & 91.4 & \cellcolor{rankfirstblue}\textbf{57.0} \\
    Qwen3.5-4B & 256K & 19.7 & 50.8 & 44.6 & 48.8 & 25.4 & 63.4 & 40.5 & 55.8 & 14.0 & 38.2 & 48.7 & 41.7 \\
    Qwen3.5-9B & 256K & 24.4 & 58.4 & 38.8 & \cellcolor{rankthirdpeach}51.8 & 30.8 & 72.1 & 38.6 & \cellcolor{rankthirdpeach}60.9 & 17.9 & 44.6 & 39.0 & \cellcolor{rankthirdpeach}42.7 \\
    Qwen3.5-27B & 256K & 26.1 & 56.1 & 92.3 & \cellcolor{ranksecondgreen}\underline{68.1} & 40.4 & 80.8 & 92.3 & \cellcolor{ranksecondgreen}\underline{84.7} & 11.8 & 31.3 & 92.3 & \cellcolor{ranksecondgreen}\underline{51.6} \\
    Gemma3-4B & 128K & 10.3 & 18.6 & 0.7 & 12.6 & 9.7 & 21.4 & 0.2 & 14.3 & 11.0 & 15.9 & 1.1 & 10.9 \\
    Gemma3-12B & 128K & 17.5 & 33.3 & 40.5 & 35.7 & 20.7 & 43.8 & 10.1 & 32.6 & 14.2 & 22.7 & 70.9 & 38.8 \\
    Gemma3-27B & 128K & 19.4 & 33.9 & 53.0 & 40.3 & 24.2 & 44.0 & 33.3 & 40.5 & 14.6 & 23.8 & 72.7 & 40.1 \\
    \rowcolor{tablegroupblue}\multicolumn{14}{c}{\rule[-0.75ex]{0pt}{2.4ex}\textit{Medical-Specialized LLMs}} \\
    MedGemma-4B & 128K & 12.2 & 22.7 & 9.8 & \cellcolor{ranksecondgreen}\underline{18.4} & 12.0 & 28.7 & 7.9 & \cellcolor{ranksecondgreen}\underline{21.8} & 12.3 & 16.7 & 11.7 & \cellcolor{ranksecondgreen}\underline{15.0} \\
    MedGemma-27B & 128K & 10.2 & 25.1 & 64.6 & \cellcolor{rankfirstblue}\textbf{38.2} & 14.9 & 37.2 & 56.4 & \cellcolor{rankfirstblue}\textbf{43.6} & 5.6 & 12.9 & 72.9 & \cellcolor{rankfirstblue}\textbf{32.9} \\
    MedMO-8B & 256K & 11.5 & 18.0 & 2.2 & 12.7 & 11.4 & 19.9 & 0.0 & 13.3 & 11.7 & 16.1 & 4.4 & 12.2 \\
    MedMO-4B & 256K & 8.0 & 12.7 & 0.4 & 8.6 & 7.5 & 14.8 & 0.0 & 9.8 & 8.4 & 10.7 & 0.9 & 7.4 \\
    MediPhi & 128K & 8.0 & 13.9 & 0.0 & 9.2 & 8.0 & 15.6 & 0.0 & 10.4 & 7.9 & 12.1 & 0.0 & 8.1 \\
    Lingshu-32B & 128K & 9.8 & 24.8 & 0.0 & \cellcolor{rankthirdpeach}16.5 & 9.4 & 27.5 & 0.0 & \cellcolor{rankthirdpeach}18.4 & 10.3 & 22.0 & 0.0 & \cellcolor{rankthirdpeach}14.7 \\
    \bottomrule
  \end{tabular}%
  }
\end{table*}

\section{Experiment Results}
\label{sec:experiment_results}

\subsection{Evaluation Protocol}
\dataset evaluates open-answer QA over longitudinal patient conversations. We report results for two evidence scopes: \emph{single-admission} questions, whose evidence is contained in one admission session, and \emph{cross-admission} questions, whose evidence spans multiple sessions. For answerable questions, we report token-level F1 and LLM-judge accuracy (J). For adversarial unanswerable questions, we report abstention accuracy (Acc), which accepts normalized equivalent abstention phrasings rather than only the stored gold wording. We also report Score, a combined metric that aggregates J on answerable questions and Acc on adversarial questions, weighted by the number of items in the corresponding split. Score is intended only as a compact relative summary for a split; it should not be read as overall model competence because high abstention can raise Score even when answer extraction remains weak. We therefore interpret F1, J, Acc, and Score together. Tables~\ref{tab:baseline_model_results} and~\ref{tab:context_method_results} report the component metrics separately, and Appendix Tables~\ref{tab:appendix_baseline_question_type_results}--\ref{tab:appendix_memory_question_type_results} provide question-type breakdowns. All table values are percentages. Appendix~\ref{app:metric_construction} gives the metric formulas, and Appendix~\ref{app:llm_judge_prompt} gives the full judge prompt.

\subsection{Models Evaluated}
\label{sec:models_evaluated}

We evaluate two groups of non-memory baselines. The general-purpose group includes GPT-5.1, Qwen3.5 models at 4B, 9B, and 27B scales, and Gemma3 models at 4B, 12B, and 27B scales. The medical/specialized group includes MedGemma-4B, MedGemma-27B, Lingshu-32B, MedMO-8B, MedMO-4B, and MediPhi. We treat MedGemma as the medical-specialized counterpart to Gemma3, which allows a direct comparison between a general model family and its medical variant.

\begin{figure*}[!t]
  \centering
  \begin{minipage}{0.49\textwidth}
    \centering
    \includegraphics[width=\linewidth]{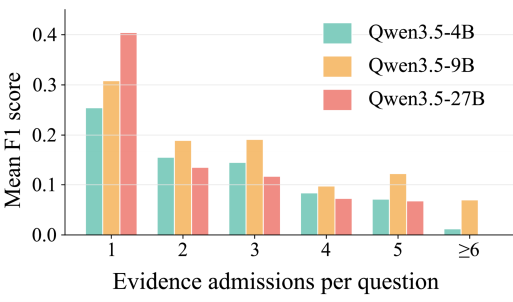}\par\vspace{2pt}
    {\small\textbf{(a) Evidence Scope Degradation}}
  \end{minipage}
  \hfill
  \begin{minipage}{0.49\textwidth}
    \centering
    \includegraphics[width=\linewidth]{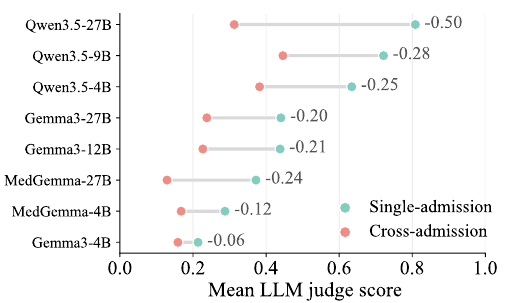}\par\vspace{2pt}
    {\small\textbf{(b) Single--Cross Admission Gap}}
  \end{minipage}
  \caption{\textbf{Answerable-question breakdowns on \dataset.} Panel (a) shows how Qwen F1 changes as supporting evidence spans more admissions, isolating the effect of evidence scope. Panel (b) compares single-admission and cross-admission LLM-judge scores for evaluated models, with labels reporting the cross-minus-single gap.}
  \label{fig:general_eval_breakdown}
\end{figure*}

\subsection{General Model Trends}
\label{sec:general_model_trends}

Among general-purpose models, Table~\ref{tab:baseline_model_results} shows that GPT-5.1 has the strongest Score across all three summary columns, driven by high judged accuracy on single-admission QA and reliable abstention. The Qwen family still exposes the benchmark's scope effect: Qwen3.5-27B has the highest single-admission F1 and strongest abstention, but its judged answer quality drops sharply once evidence spans multiple admissions, while Qwen3.5-9B is stronger on cross-admission answerable metrics. This explains why Score should be read with its components rather than as a standalone ranking.

Gemma3 improves with scale mainly by abstaining more reliably. Its Score rises partly because Acc improves, but even the largest Gemma3 model remains below Qwen on judge-based answer quality, separating better abstention from stronger longitudinal evidence extraction. Appendix~\ref{app:question_type_results} provides full question-type breakdowns.

\subsection{Medical and Specialized Models}
\label{sec:medical_specialized_models}

The medical/specialized models do not consistently outperform the strongest general-purpose baselines. MedGemma-27B has the highest group-level Score and Acc, but this mainly reflects abstention and local behavior rather than stronger cross-admission answer extraction. On cross-admission answerable questions, it remains weak despite strong adversarial abstention accuracy.

Other specialized models show complementary weaknesses. MedGemma-4B has the best specialized-model F1 overall and on cross-admission questions, while Lingshu-32B has the highest specialized-model cross-admission J but fails on adversarial abstention. MedMO and MediPhi remain low across both answer quality and abstention, suggesting that domain specialization alone does not resolve the benchmark's longitudinal evidence-use challenge in this model set.

\begin{table*}[!t]
  \centering
  \small
  \setlength{\tabcolsep}{3pt}
  \caption{\textbf{Memory-augmented results for Gemma3 models on \dataset, with BM25 and Embedding included as additional baseline results.} The table uses the same percentage metrics as the main baseline evaluation: F1 and J for answerable questions, and Acc for adversarial unanswerable questions. Score is the item-weighted combination of J on answerable items and Acc on adversarial items within the corresponding split. In Score columns, \protect\colorbox{rankfirstblue}{\protect\textbf{first}}, \protect\colorbox{ranksecondgreen}{\protect\underline{second}}, and \protect\colorbox{rankthirdpeach}{third} mark summary ranks across rows.}
  \label{tab:context_method_results}
  \resizebox{\textwidth}{!}{%
  \begin{tabular}{@{}l@{\hspace{4pt}}@{\hspace{4pt}}l@{\hspace{18pt}}cccc@{\hspace{18pt}}cccc@{\hspace{18pt}}cccc@{}}
    \toprule
    \multicolumn{1}{@{}l}{\smash{\raisebox{-1.15\normalbaselineskip}{\shortstack[l]{\textbf{Context}\\\textbf{Method}}}}} & \multicolumn{1}{@{\hspace{4pt}}l@{\hspace{18pt}}}{\smash{\raisebox{-0.7\normalbaselineskip}{\textbf{Model}}}} & \multicolumn{4}{c}{\textbf{Overall}} & \multicolumn{4}{c}{\textbf{Single-admission}} & \multicolumn{4}{c}{\textbf{Cross-admission}} \\
    \cmidrule(lr){3-6}\cmidrule(lr){7-10}\cmidrule(l){11-14}
    & & F1 & J & Acc & Score & F1 & J & Acc & Score & F1 & J & Acc & Score \\
    \midrule
    mem0 & Gemma3-4B & 16.3 & 32.3 & 20.2 & 28.2 & 17.3 & 37.8 & 10.6 & 28.7 & 15.3 & 26.8 & 29.7 & 27.8 \\
    mem0 & Gemma3-12B & 19.8 & 42.6 & 65.0 & \cellcolor{rankthirdpeach}50.1 & 25.4 & 52.1 & 49.6 & 51.2 & 14.2 & 33.1 & 80.4 & 48.9 \\
    mem0 & Gemma3-27B & 18.3 & 42.6 & 87.5 & \cellcolor{rankfirstblue}\textbf{57.6} & 25.4 & 52.5 & 87.7 & \cellcolor{ranksecondgreen}\underline{64.2} & 11.3 & 32.6 & 87.5 & \cellcolor{rankfirstblue}\textbf{50.9} \\
    \midrule
    mem0* & Gemma3-4B & 13.1 & 22.2 & 23.8 & 22.8 & 12.7 & 23.1 & 11.7 & 19.3 & 13.5 & 21.4 & 35.9 & 26.2 \\
    mem0* & Gemma3-12B & 16.9 & 35.2 & 66.6 & 45.7 & 21.1 & 39.3 & 50.2 & 43.0 & 12.8 & 31.2 & 83.0 & 48.5 \\
    mem0* & Gemma3-27B & 18.3 & 41.1 & 90.6 & \cellcolor{rankfirstblue}\textbf{57.6} & 24.9 & 52.0 & 91.4 & \cellcolor{rankfirstblue}\textbf{65.1} & 11.7 & 30.3 & 89.9 & \cellcolor{ranksecondgreen}\underline{50.1} \\
    \midrule
    mem$\alpha$ & Gemma3-4B & 11.0 & 12.6 & 24.8 & 16.6 & 10.1 & 12.6 & 20.9 & 15.3 & 11.9 & 12.6 & 28.6 & 17.9 \\
    mem$\alpha$ & Gemma3-12B & 14.3 & 27.8 & 72.0 & 42.5 & 16.5 & 29.7 & 62.5 & 40.7 & 12.0 & 25.8 & 81.5 & 44.4 \\
    mem$\alpha$ & Gemma3-27B & 14.7 & 33.5 & 89.4 & \cellcolor{ranksecondgreen}\underline{52.1} & 18.0 & 38.0 & 89.0 & \cellcolor{rankthirdpeach}55.0 & 11.4 & 29.0 & 89.9 & \cellcolor{rankthirdpeach}49.3 \\
    \midrule
    BM25 & Gemma3-4B & 12.9 & 26.9 & 0.8 & 18.2 & 14.8 & 35.5 & 0.2 & 23.7 & 11.0 & 18.3 & 1.3 & 12.6 \\
    BM25 & Gemma3-12B & 20.4 & 40.1 & 37.4 & 39.2 & 25.9 & 53.2 & 8.6 & 38.3 & 14.8 & 27.0 & 66.3 & 40.1 \\
    BM25 & Gemma3-27B & 21.2 & 41.3 & 52.2 & 44.9 & 27.8 & 55.1 & 33.5 & 47.9 & 14.6 & 27.5 & 70.9 & 42.0 \\
    \midrule
    Embedding & Gemma3-4B & 12.2 & 23.6 & 1.1 & 16.1 & 13.3 & 29.6 & 0.4 & 19.9 & 11.2 & 17.6 & 1.8 & 12.3 \\
    Embedding & Gemma3-12B & 18.7 & 37.2 & 38.8 & 37.7 & 23.0 & 47.4 & 8.6 & 34.4 & 14.3 & 27.0 & 68.9 & 41.0 \\
    Embedding & Gemma3-27B & 19.8 & 37.5 & 56.0 & 43.6 & 24.5 & 46.7 & 39.4 & 44.3 & 15.1 & 28.2 & 72.5 & 43.0 \\
    \bottomrule
  \end{tabular}%
  }
\end{table*}

\subsection{Memory-Augmented Evaluation}
\label{sec:memory}

\paragraph{Motivation.}
We focus memory augmentation on Gemma3 because its 128K token context is the main setting where \dataset can exceed the usable full-history prompt budget. Qwen3.5 models provide longer 256K token windows, while the longest \dataset timelines can approach or exceed the Gemma3 budget once questions and evaluation instructions are added. The Gemma3 setting therefore tests whether external memory can preserve useful longitudinal evidence when direct full-context prompting would otherwise require truncating the full patient history.

\paragraph{Memory methods and retrieval baselines.}
We evaluate three memory methods with Gemma3 models. \textit{mem0} follows the long-term memory framework of Mem0~\citep{chhikara2025mem0}, which maintains an external memory store while sequentially processing conversation history. \textit{mem0*} uses the same overall memory interface but restricts the policy to write-only memory additions, and \textit{mem$\alpha$} uses the multi-component memory structure introduced by Mem-\(\alpha\)~\citep{wang2025memalpha}. To contextualize the memory results, we also include additional baseline results from BM25~\citep{robertson2009probabilistic} and Embedding retrieval based on retrieval-augmented generation~\citep{lewis2020retrieval}. Appendix~\ref{app:context_method_details} summarizes the evaluated memory and retrieval methods.

\paragraph{Results.}
Table~\ref{tab:context_method_results} shows that memory augmentation improves Gemma3 mainly through better abstention and, for larger models, stronger judged answer quality. The Score summaries also rise, but much of that movement comes from Acc, so the component metrics are essential for interpretation. The retrieval baselines provide a useful check on whether the gains come only from shortening the input: they improve over plain Gemma3, but remain below the strongest memory variants. In this Gemma3 augmentation study, the remaining gap suggests that \dataset stresses evidence selection and cross-admission linking, not only context-window length. Among memory methods, the highlighted summary cells favor mem0 and mem0*, while the component metrics show a mixed pattern: mem0* is stronger for single-admission behavior, and mem0 is more favorable on cross-admission questions. The more structured mem$\alpha$ variant remains competitive but does not dominate, indicating that richer memory organization is not automatically beneficial without task alignment. Appendix~\ref{app:question_type_results} provides the full question-type breakdown table results.

\paragraph{Implications for memory design.}
The main design implication is that memory should support evidence selection and cross-session linking, not just compress long histories into shorter prompts. Score is useful as a summary, but it should be read with F1/J/Acc and question-type results because stronger abstention can raise Score while answer extraction remains weak. Constrained variants such as mem0* and simple retrieval baselines are therefore important ablations: they test whether added memory actions and structure improve benchmark behavior beyond simpler context selection. Richer structures such as mem$\alpha$ may still be useful, but the results suggest they need task-aligned training or tuning to translate structure into stronger cross-admission clinical reasoning.

\subsection{Unanswerable Behavior and Cross-Admission Difficulty}
\label{sec:unanswerable_cross_admission}

Adversarial unanswerable performance is highly uneven. Qwen3.5-27B is the strongest abstainer across evidence scopes, and larger Gemma3 and MedGemma variants also improve on this dimension. However, abstention does not imply answer extraction: MedGemma-27B leads specialized models on Acc while trailing other specialized models on cross-admission answer extraction.

Figure~\ref{fig:general_eval_breakdown} shows why cross-admission QA is the central stress test in \dataset. Qwen F1 drops as evidence spans more admissions, especially for Qwen3.5-27B: the strongest single-admission model loses its advantage as scope expands. Every displayed model also has lower cross-admission J than single-admission J, with the largest gap again for Qwen3.5-27B. Together, Table~\ref{tab:baseline_model_results} and Figure~\ref{fig:general_eval_breakdown} show that the evaluated models use local evidence or abstain more reliably than they ground answers in evidence distributed across encounters.

This separation matters because a useful long-context clinical system should extract local facts, connect them across encounters, and reject unsupported questions without letting one behavior substitute for the others.

\section{Conclusion}
\label{sec:conclusion}

We introduced \dataset, a patient-level benchmark for evaluating long-context LLMs on longitudinal clinical reasoning. Unlike short-context medical QA or aggregate long-context scores, \dataset turns multi-admission EHR-derived timelines into evidence-linked questions that separately test local evidence use, cross-admission integration, and principled abstention when the record does not support an answer. This structure makes the benchmark useful for diagnosing where a system fails: whether it loses relevant details in long inputs, cannot connect evidence across encounters, or improves apparent performance by abstaining more often. In the evaluated suite, larger context windows, medical specialization, retrieval baselines, and external memory each help in specific ways, but none eliminates the gap between local evidence use and cross-admission reasoning. \dataset therefore provides a focused testbed for measuring whether long-context clinical systems can reliably link evidence across patient timelines.


\section*{Limitations}

\dataset has several limitations. First, it uses a text-only design: generated doctor--patient conversations do not include clinical images or other non-text media, so radiology evidence is represented through reports rather than image pixels. Second, the conversations and QA items are synthetic transformations of deidentified EHR evidence. This makes large-scale, evidence-linked evaluation possible, but it may introduce artifacts in patient phrasing, clinical dialogue flow, and question style that differ from real patient--clinician interactions. Third, the benchmark intentionally selects patients with many eligible admissions to stress long-context reasoning, which can overrepresent admission-rich trajectories and underrepresent shorter or less documented care histories. Future work should develop \dataset beyond this text-only setting, add clinician-facing audits of generated dialogue and QA, and broaden patient sampling while preserving the benchmark's focus on longitudinal evidence use.

\section*{Ethics Statement}

\dataset is intended as a research benchmark for evaluating long-context clinical reasoning, not as a clinical decision-support system or patient-facing tool. The released \dataset benchmark contains synthetic doctor--patient conversations, summaries, QA items, and evaluation metadata, and is downloadable for research review; it does not redistribute raw MIMIC-IV or MIMIC-IV-Note tables or notes. In the code release, the folders reserved for MIMIC-IV and MIMIC-IV-Note inputs are empty placeholders. Users who want to rerun the construction pipeline must obtain credentialed access to the underlying MIMIC resources, comply with their data-use requirements, and place the authorized files locally in those folders. Although the source records are deidentified and the released conversations are synthetic, the benchmark is derived from real clinical documentation; users should treat it as sensitive medical data, avoid reidentification attempts, and interpret model outputs as benchmark behavior rather than clinical advice.


\bibliography{references}

\appendix
\section{Details of \dataset}
\label{app:medlocomo_details}

\dataset is released as a patient-level benchmark directory rather than as a single flat file. This appendix describes the released artifact layout and the fields needed to run or audit evaluation. Section~\ref{sec:dataset_generation} describes how the artifacts are generated, and Appendix~\ref{app:generation_prompts} provides detailed prompt templates and concrete examples.

\paragraph{Generation model.}
Conversation, admission-summary, and QA artifacts are generated with Qwen3-235B-A22B-Instruct-2507.\footnote{\href{https://huggingface.co/Qwen/Qwen3-235B-A22B-Instruct-2507}{Hugging Face model page}.} This model is used only for benchmark construction. It is separate from the evaluated models and belongs to a different Qwen model family than the Qwen3.5 baselines used throughout the main evaluation suite.

\subsection{Generation Prompt Structure}

The conversation-generation prompt presents clinical evidence in three progressively richer blocks: a discharge-note storyline, an ordered problem list derived from ICD diagnosis titles, and a supporting-context block containing truncated radiology notes and selected procedure and microbiology entries. The implementation includes up to 8 radiology notes and up to 10 procedure and microbiology entries to keep prompts bounded while preserving diverse clinical evidence. The full raw packet is included only as backup context. The prompt also provides a recommended turn-count range based on length of stay, allowing longer admissions to produce richer conversations while keeping outputs predictable. Appendix~\ref{app:generation_prompts} gives the full generation prompt templates and examples.

\subsection{Released Dataset Layout}

The benchmark release is organized hierarchically. The benchmark root contains one folder per patient, named by patient identifier. Each patient folder contains patient-level files and one subfolder per processed admission. Admission folders are named by admission identifier and contain the generated conversation, admission summary, admission-level QA, and audit artifacts for that admission. At a high level, the release contains:
\begin{itemize}
    \item Patient-level evaluation files: combined conversation, patient summary, benchmark QA, and cross-admission QA.
    \item Admission-level evaluation files: conversation, summary, and QA.
    \item Admission-level audit files: formed packet and prompt record, which support traceability but are not required for model evaluation.
\end{itemize}

\subsection{Patient-Level Files}

The combined conversation file stores the full patient timeline as a chronological list of admission sessions. Each session records the admission identifier, admission start and end times, and generated conversation lines. This is the primary long-context input for full-history model evaluation.

The patient summary file records patient-level metadata, including eligible and processed admission counts, processed admission identifiers, LLM call statistics, total and mean conversation turns, total and mean token counts, and tokenizer name. The benchmark QA file contains the shuffled final QA set for that patient, while the cross-admission QA file keeps the cross-session subset separate for split-level analysis.

\subsection{Admission-Level Files}

Each admission subfolder contains the generated conversation, summary, and QA for one hospital stay. The conversation consists of ordered turns with turn number, timestamp, speaker, and text. The summary contains the admission window, a one-paragraph hospital-course summary, and a problem list. The admission-level QA file contains the two answerable single-admission questions and one adversarial unanswerable question generated for the target admission.

The formed packet and prompt record are included for reproducibility and auditing. They are not required for model evaluation, but they make it possible to trace how a generated admission session was constructed from the underlying evidence and how the generation request was issued.

\subsection{Timestamp Normalization}
\label{app:timestamp_normalization}

MIMIC contains both precise timestamps and date-only fields. We normalize times to make ordering deterministic and validation reproducible. Date-only procedure and microbiology events are normalized to noon. Diagnoses often lack reliable within-stay timing, so we assign pseudo-timestamps near discharge, anchored to the end of the stay. If discharge time is unavailable, admission end falls back to the latest discharge-note chart time or store time. These rules preserve temporal order without claiming unsupported precision. Accordingly, QA construction avoids fine-grained within-stay temporal comparisons whose answer would depend on normalized pseudo-times; temporal questions rely on admission order or explicit conversation evidence in the dialogue.

\subsection{QA and Evidence Format}

Each QA item contains id, scope, question type, question, answer, and evidence. Single-admission answerable questions cover medical reasoning and care plan rationale; cross-admission questions cover longitudinal progression, cross-admission comparison, and frequency pattern. Evidence cites supporting turns for single-admission items and the required admissions for cross-admission items. Adversarial items use plausible but unsupported questions with the stored gold answer ``the question is not answerable.''

Adversarial questions are designed to be close to regular answerable questions rather than visibly out of scope. They preserve the clinical topic and reasoning frame while minimally changing the question so the record no longer supports an answer, forcing models to distinguish semantic similarity from actual evidence support.

\subsection{Benchmark QA Statistics}
\label{app:benchmark_qa_statistics}

The full benchmark contains 17,892 QA items across 100 patients, or 178.9 questions per patient on average. Of these, 11,928 are answerable and 5,964 are adversarial unanswerable. Single-admission QA contributes 8,946 items, and cross-admission QA contributes another 8,946 items. Table~\ref{tab:appendix_evidence_span_distribution} reports the benchmark-level evidence-span distribution. The one-admission row corresponds to single-admission QA, while the remaining rows quantify how cross-admission questions are distributed across two, three, or at least four admissions in total.

\begin{table}[t]
  \centering
  \small
  \setlength{\tabcolsep}{4pt}
  \caption{\textbf{Evidence-admission span distribution in \dataset.} Percent of all QA is computed over all 17,892 questions; percent of cross-admission QA is computed over the 8,946 questions whose evidence spans multiple patient admissions.}
  \label{tab:appendix_evidence_span_distribution}
  \resizebox{\columnwidth}{!}{%
  \begin{tabular}{lrrr}
    \toprule
    Evidence admissions & Questions & All QA & Cross QA \\
    \midrule
    1 & 8,946 & 50.00\% & -- \\
    2 & 5,282 & 29.52\% & 59.04\% \\
    3 & 2,497 & 13.96\% & 27.91\% \\
    4+ & 1,167 & 6.52\% & 13.04\% \\
    \bottomrule
  \end{tabular}%
  }
\end{table}

\subsection{Quality Control Checks}

Because the full benchmark is too large for complete manual inspection, the sampled human audit in Section~\ref{sec:dataset_generation} is complemented by automated checks over the released artifacts. The generation pipeline applies these checks before patient-level artifacts are released. Conversation checks enforce grounded clinical facts from the EHR-derived packet, valid schema, exact speaker labels, contiguous turns, nonempty text, monotonic timestamps, and admission-window consistency. QA checks validate each answer with its cited evidence: single-admission items must cite supporting turns from the target admission, cross-admission items must cite valid patient admissions, and adversarial items must use the standard abstention gold answer in the released QA file. QA construction also avoids fine-grained within-stay temporal comparisons that would depend on normalized pseudo-times. Together, these checks keep generated dialogue and evidence-linked answers traceable to the admission packets.

\section{Additional Experimental Details and Results}
\label{app:additional_results}

\subsection{Metric Construction}
\label{app:metric_construction}

Let \(S\) denote an evaluation split such as overall, single-admission, or cross-admission. We write \(A_S\) for the answerable questions in the split and \(U_S\) for adversarial unanswerable questions. All table values are reported as percentages in every reported result table.

\paragraph{Answerable F1.}
For each answerable question \(i\), the prediction \(\hat{a}_i\) and gold answer \(a_i\) are normalized by lowercasing, removing punctuation, removing articles, and collapsing whitespace. Let \(T(\cdot)\) be the resulting multiset of tokens. Whole-answer precision, recall, and F1 are:
\[
P_i=\frac{|T(\hat{a}_i)\cap T(a_i)|}{|T(\hat{a}_i)|},
\quad
R_i=\frac{|T(\hat{a}_i)\cap T(a_i)|}{|T(a_i)|},
\]
\[
F_i^{\mathrm{whole}} =
\begin{cases}
\frac{2P_iR_i}{P_i+R_i}, & P_i+R_i>0,\\
0, & P_i+R_i=0.
\end{cases}
\]
When either answer contains comma-separated spans, the evaluator also computes a comma-aware score. For prediction parts \(\{\hat{p}_{ik}\}\) and gold parts \(\{g_{i\ell}\}\),
\[
P^{\mathrm{comma}}_i =
\frac{1}{K}\sum_k \max_\ell F^{\mathrm{whole}}(\hat{p}_{ik}, g_{i\ell}),
\]
\[
R^{\mathrm{comma}}_i =
\frac{1}{L}\sum_\ell \max_k F^{\mathrm{whole}}(\hat{p}_{ik}, g_{i\ell}),
\]
and \(F_i^{\mathrm{comma}}\) is the harmonic mean of \(P^{\mathrm{comma}}_i\) and \(R^{\mathrm{comma}}_i\). The reported per-question F1 is
\[
F1_i = \max(F_i^{\mathrm{whole}}, F_i^{\mathrm{comma}}),
\]
with \(F_i^{\mathrm{whole}}\) used when no comma-aware score is available. Split-level F1 is the mean over answerable questions:
\[
F1_S=\frac{1}{|A_S|}\sum_{i\in A_S}F1_i.
\]

\paragraph{LLM-judge accuracy.}
For each answerable question, the judge returns \(j_i\in\{0,1\}\). The LLM-judge score reported as \(J\) is
\[
J_S=\frac{1}{|A_S|}\sum_{i\in A_S}j_i.
\]

\paragraph{Adversarial abstention accuracy.}
Adversarial gold answers use the canonical phrase ``the question is not answerable,'' but scoring is not exact-string. Let \(\mathcal{C}_{\mathrm{abs}}\) be the accepted set of normalized abstention phrasings, including common equivalents such as ``not answerable from the record,'' ``cannot be determined,'' and ``not mentioned.'' After case folding, punctuation removal, article removal, and whitespace normalization,
\[
\mathrm{Acc}_i =
\mathbf{1}\{\mathrm{norm}(\hat{a}_i)\in\mathcal{C}_{\mathrm{abs}}\}.
\]
The split-level adversarial accuracy is
\[
\mathrm{Acc}_S=\frac{1}{|U_S|}\sum_{i\in U_S}\mathrm{Acc}_i.
\]

\paragraph{Combined Score.}
The combined Score uses LLM-judge accuracy for answerable questions and abstention accuracy for adversarial questions:
\[
\mathrm{Score}_S =
\frac{\sum_{i\in A_S}j_i+\sum_{i\in U_S}\mathrm{Acc}_i}
{|A_S|+|U_S|}.
\]

\subsection{LLM Judge Prompt}
\label{app:llm_judge_prompt}

For answerable predictions, we grade all candidate answers with a single fixed LLM judge, Gemini 3 Flash Preview, using the model code \texttt{gemini-3-flash-preview}.\footnote{\href{https://ai.google.dev/gemini-api/docs/models/gemini-3-flash-preview}{Google AI for Developers model documentation}.} Before final evaluation, we tested several judge choices and found completely consistent binary decisions together with human review, which is expected because the task is a constrained 0/1 correctness judgment. We therefore use one fixed judge for consistency across all reported results. The judge is used only for binary grading after model answers are produced; it is not used to generate \dataset or to answer benchmark questions. Its decisions were checked by human review for consistency with the grading rubric. Adversarial questions are scored directly by the abstention-phrase matcher above rather than by the judge. The judge prompt structure is:

\begin{promptbox}{Answerable QA Judge Prompt}
=== System Message ===
You are grading candidate answers for short-answer
medical benchmark questions.

Judge only from the provided question, gold_answer,
and candidate_answer.

Score 1 when the candidate answer is correct.

Score 0 when the candidate answer is false, incorrect,
unsupported, incomplete enough to be wrong, or only
says it is not answerable.

Return strict JSON with the schema
{"judgments": [{"qa_id": "...", "score": 1}]}.

Each score must be exactly one of: 0, 1.
Return exactly one judgment per provided qa_id.

=== User Message ===
{
  "items": [
    {
      "qa_id": "...",
      "question": "...",
      "gold_answer": "...",
      "candidate_answer": "..."
    }
  ]
}
\end{promptbox}

\subsection{Question-Type Results}
\label{app:question_type_results}

Tables~\ref{tab:appendix_baseline_question_type_results} and~\ref{tab:appendix_memory_question_type_results} break down results by question type. For answerable regular question types, each block reports token-level F1 and LLM-judge accuracy (J). For adversarial question types, the tables report abstention accuracy (Acc). All values are reported as one-decimal percentage values.

Table~\ref{tab:appendix_baseline_question_type_results} shows that GPT-5.1 leads the local answerable judge scores among general-purpose models and is also strong on adversarial abstention. Qwen models remain strongest on several longitudinal question types: Qwen3.5-9B leads longitudinal progression, cross-admission comparison, and frequency pattern, while Qwen3.5-27B leads adversarial abstention. Medical-specialized models are weaker overall, but Lingshu-32B is comparatively strong on longitudinal and cross-admission type judge scores, while MedGemma-27B is the strongest medical model on adversarial abstention accuracy.

Table~\ref{tab:appendix_memory_question_type_results} shows that memory augmentation helps most clearly through adversarial Acc and selected Gemma3 answerable J scores, with BM25 and Embedding included as additional baseline results. The strongest memory cells are concentrated in mem0 and mem0*, while mem$\alpha$ remains competitive for Gemma3-27B on adversarial Acc but does not consistently dominate. The retrieval baselines are competitive on some local or pattern-oriented judge scores, but they remain weaker on the adversarial and longitudinal columns where the best memory methods are strongest, reinforcing that the benchmark stresses cross-admission evidence selection and linking rather than only context-window length by itself.

\subsection{Memory Methods and Retrieval Baselines}
\label{app:context_method_details}

\paragraph{Memory construction.}
The memory-augmented experiments use the same QA evaluation protocol and metrics as the main baseline results. For each patient, the conversation is processed chronologically by admission-local chunks. Each chunk is passed to a memory-construction call together with a compact rolling patient summary and a small window of recent chunk summaries. The rolling summary is capped at 1,200 characters and is used only to preserve longitudinal context for later memory extraction; answer prompts are built from retrieved memories rather than rereading the rolling patient summary during answer generation.

\paragraph{Store size and update policy.}
The memory store is data-dependent rather than fixed to a prespecified final size. Each chunk can propose only a bounded number of candidate memories, but the final store may grow or shrink depending on the update policy; every run logs total, active, deleted, duplicate-skipped, added, updated, and no-op memory counts. In \textit{mem0}, the extractor first proposes atomic timestamp-aware memories. A dense local retriever then finds similar active memories, and an update call chooses among ADD, UPDATE, DELETE, and NOOP actions. Invalid or missing update decisions fall back to ADD, while DELETE and UPDATE must target an existing active memory. In \textit{mem0*}, the same extraction interface is used but the policy is write-only: nonduplicate candidates are appended as new memories and exact normalized duplicates are skipped, so there are no delete or merge decisions. In \textit{mem$\alpha$}, extracted memories are also add-only but are typed into core, episodic, and semantic stores following the Mem-\(\alpha\) organization~\citep{wang2025memalpha}. This separates durable patient facts, timestamped events, and reusable patient-specific clinical relations while keeping the same downstream evaluation pipeline.

\paragraph{Answer-time retrieval and budgets.}
For each benchmark question, memory methods query the active memory store using the question text. The dense retriever returns the top ranked memories, the answer prompt renders at most the configured maximum number of memories, and retrieved memories are shown in chronological order when visible timestamps are available. If the estimated prompt exceeds the effective budget, defined by model context length minus answer-generation tokens and a safety margin, the lowest-ranked memories are pruned until the answer prompt fits.

\paragraph{Retrieval baselines.}
The retrieval baselines use the same Gemma3 backbones but do not maintain a memory store. They build one document per admission from the generated doctor--patient conversation, score all admissions for each question, render the selected admissions chronologically, and prune lower-ranked admissions if the prompt exceeds the same token budget. BM25 uses lexical retrieval with \(k_1=1.5\) and \(b=0.75\)~\citep{robertson2009probabilistic}. Embedding uses the same local dense Hugging Face retrieval stack as the memory retriever in a retrieval-augmented generation setup~\citep{lewis2020retrieval}.

\begin{table*}[!t]
  \centering
  \small
  \setlength{\tabcolsep}{3pt}
  \caption{\textbf{Baseline question-type results on \dataset.} MR = medical reasoning, CPR = care plan rationale, LP = longitudinal progression, CAC = cross-admission comparison, FP = frequency pattern, Single Adv. = single-admission adversarial, and Cross Adv. = cross-admission adversarial. For regular question types, F1 and J evaluate answerable QA; for adversarial question types, Acc evaluates abstention. Colored J and Acc cells mark top-three ranks within each model group: \protect\colorbox{rankfirstblue}{\protect\textbf{first}}, \protect\colorbox{ranksecondgreen}{\protect\underline{second}}, and \protect\colorbox{rankthirdpeach}{third place}.}
  \label{tab:appendix_baseline_question_type_results}
  \resizebox{\textwidth}{!}{%
  \begin{tabular}{@{}l@{\hspace{4pt}}@{\hspace{4pt}}l@{\hspace{15pt}}cc@{\hspace{15pt}}cc@{\hspace{15pt}}cc@{\hspace{15pt}}cc@{\hspace{15pt}}cc@{\hspace{15pt}}c@{\hspace{15pt}}c@{}}
    \toprule
    \multicolumn{1}{@{}l}{\smash{\raisebox{-0.7\normalbaselineskip}{\textbf{Model}}}} & \multicolumn{1}{@{\hspace{1pt}}c@{\hspace{8pt}}}{\smash{\raisebox{-1.15\normalbaselineskip}{\shortstack[c]{\textbf{Context}\\\textbf{Length}}}}} & \multicolumn{2}{c}{\textbf{MR}} & \multicolumn{2}{c}{\textbf{CPR}} & \multicolumn{2}{c}{\textbf{LP}} & \multicolumn{2}{c}{\textbf{CAC}} & \multicolumn{2}{c}{\textbf{FP}} & \multicolumn{1}{c}{\textbf{Single Adv.}} & \multicolumn{1}{c}{\textbf{Cross Adv.}} \\
    \cmidrule(lr){3-4}\cmidrule(lr){5-6}\cmidrule(lr){7-8}\cmidrule(lr){9-10}\cmidrule(lr){11-12}\cmidrule(lr){13-13}\cmidrule(l){14-14}
    & & F1 & J & F1 & J & F1 & J & F1 & J & F1 & J & Acc & Acc \\
    \midrule
    \rowcolor{tablegroupblue}\multicolumn{14}{c}{\rule[-0.75ex]{0pt}{2.4ex}\textit{General-Purpose LLMs}} \\
    GPT-5.1 & 400K & 31.2 & \cellcolor{rankfirstblue}\textbf{87.2} & 25.2 & \cellcolor{rankfirstblue}\textbf{89.7} & 16.8 & \cellcolor{rankthirdpeach}45.1 & 9.8 & \cellcolor{ranksecondgreen}\underline{35.7} & 13.3 & \cellcolor{ranksecondgreen}\underline{34.1} & \cellcolor{ranksecondgreen}\underline{89.0} & \cellcolor{ranksecondgreen}\underline{91.4} \\
    Qwen3.5-4B & 256K & 27.1 & 61.7 & 23.8 & 65.2 & 18.9 & \cellcolor{ranksecondgreen}\underline{46.5} & 10.2 & \cellcolor{rankthirdpeach}31.5 & 8.3 & \cellcolor{rankthirdpeach}29.3 & \cellcolor{rankthirdpeach}40.5 & 48.7 \\
    Qwen3.5-9B & 256K & 34.3 & \cellcolor{rankthirdpeach}71.6 & 27.4 & \cellcolor{rankthirdpeach}72.7 & 22.4 & \cellcolor{rankfirstblue}\textbf{51.5} & 12.8 & \cellcolor{rankfirstblue}\textbf{41.2} & 14.6 & \cellcolor{rankfirstblue}\textbf{34.9} & 38.6 & 39.0 \\
    Qwen3.5-27B & 256K & 45.2 & \cellcolor{ranksecondgreen}\underline{80.8} & 35.7 & \cellcolor{ranksecondgreen}\underline{80.8} & 15.7 & 36.3 & 6.8 & 25.6 & 9.7 & 27.5 & \cellcolor{rankfirstblue}\textbf{92.3} & \cellcolor{rankfirstblue}\textbf{92.3} \\
    Gemma3-4B & 128K & 9.0 & 18.9 & 10.3 & 23.8 & 15.1 & 22.2 & 8.3 & 7.1 & 5.8 & 12.7 & 0.2 & 1.1 \\
    Gemma3-12B & 128K & 22.2 & 40.3 & 19.1 & 47.4 & 17.4 & 26.8 & 8.5 & 10.9 & 14.1 & 27.1 & 10.1 & 70.9 \\
    Gemma3-27B & 128K & 23.9 & 41.4 & 24.5 & 46.7 & 18.6 & 30.2 & 8.5 & 11.8 & 13.1 & 24.0 & 33.3 & \cellcolor{rankthirdpeach}72.7 \\
    \rowcolor{tablegroupblue}\multicolumn{14}{c}{\rule[-0.75ex]{0pt}{2.4ex}\textit{Medical-Specialized LLMs}} \\
    MedGemma-4B & 128K & 12.0 & \cellcolor{ranksecondgreen}\underline{29.5} & 12.0 & \cellcolor{ranksecondgreen}\underline{28.0} & 16.4 & \cellcolor{ranksecondgreen}\underline{21.8} & 10.2 & \cellcolor{rankthirdpeach}11.8 & 6.6 & \cellcolor{rankthirdpeach}12.2 & \cellcolor{ranksecondgreen}\underline{7.9} & \cellcolor{ranksecondgreen}\underline{11.7} \\
    MedGemma-27B & 128K & 15.2 & \cellcolor{rankfirstblue}\textbf{36.8} & 14.6 & \cellcolor{rankfirstblue}\textbf{37.7} & 7.1 & 16.3 & 5.1 & 9.2 & 3.1 & 10.0 & \cellcolor{rankfirstblue}\textbf{56.4} & \cellcolor{rankfirstblue}\textbf{72.9} \\
    MedMO-8B & 256K & 11.1 & 18.5 & 11.6 & 21.4 & 14.8 & \cellcolor{rankthirdpeach}18.1 & 9.3 & \cellcolor{ranksecondgreen}\underline{13.0} & 8.1 & \cellcolor{ranksecondgreen}\underline{15.3} & 0.0 & \cellcolor{rankthirdpeach}4.4 \\
    MedMO-4B & 256K & 6.2 & 15.0 & 8.8 & 14.5 & 11.0 & 14.1 & 7.8 & 7.1 & 4.0 & 7.9 & 0.0 & 0.9 \\
    MediPhi & 128K & 7.8 & 14.8 & 8.2 & 16.5 & 9.7 & 13.2 & 6.5 & 11.3 & 5.9 & 10.9 & 0.0 & 0.0 \\
    Lingshu-32B & 128K & 9.0 & \cellcolor{rankthirdpeach}28.2 & 9.8 & \cellcolor{rankthirdpeach}26.9 & 13.6 & \cellcolor{rankfirstblue}\textbf{25.4} & 8.3 & \cellcolor{rankfirstblue}\textbf{18.5} & 6.2 & \cellcolor{rankfirstblue}\textbf{19.2} & 0.0 & 0.0 \\
    \bottomrule
  \end{tabular}%
  }
\end{table*}

\begin{table*}[!t]
  \centering
  \small
  \setlength{\tabcolsep}{3pt}
  \caption{\textbf{Memory-augmented question-type results on \dataset, with BM25 and Embedding included as additional baseline results.} MR = medical reasoning, CPR = care plan rationale, LP = longitudinal progression, CAC = cross-admission comparison, FP = frequency pattern, Single Adv. = single-admission adversarial, and Cross Adv. = cross-admission adversarial. For regular question types, F1 and J evaluate answerable QA; for adversarial question types, Acc evaluates abstention. Colored J and Acc cells mark top-three ranks across rows: \protect\colorbox{rankfirstblue}{\protect\textbf{first}}, \protect\colorbox{ranksecondgreen}{\protect\underline{second}}, and \protect\colorbox{rankthirdpeach}{third place}.}
  \label{tab:appendix_memory_question_type_results}
  \resizebox{\textwidth}{!}{%
  \begin{tabular}{@{}l@{\hspace{4pt}}c@{\hspace{15pt}}cc@{\hspace{15pt}}cc@{\hspace{15pt}}cc@{\hspace{15pt}}cc@{\hspace{15pt}}cc@{\hspace{15pt}}c@{\hspace{15pt}}c@{}}
    \toprule
    \multicolumn{1}{@{}l}{\smash{\raisebox{-1.15\normalbaselineskip}{\shortstack[l]{\textbf{Context}\\\textbf{Method}}}}} & \multicolumn{1}{@{\hspace{4pt}}l@{\hspace{15pt}}}{\smash{\raisebox{-0.7\normalbaselineskip}{\textbf{Model}}}} & \multicolumn{2}{c}{\textbf{MR}} & \multicolumn{2}{c}{\textbf{CPR}} & \multicolumn{2}{c}{\textbf{LP}} & \multicolumn{2}{c}{\textbf{CAC}} & \multicolumn{2}{c}{\textbf{FP}} & \multicolumn{1}{c}{\textbf{Single Adv.}} & \multicolumn{1}{c}{\textbf{Cross Adv.}} \\
    \cmidrule(lr){3-4}\cmidrule(lr){5-6}\cmidrule(lr){7-8}\cmidrule(lr){9-10}\cmidrule(lr){11-12}\cmidrule(lr){13-13}\cmidrule(l){14-14}
    & & F1 & J & F1 & J & F1 & J & F1 & J & F1 & J & Acc & Acc \\
    \midrule
    mem0 & Gemma3-4B & 20.1 & 42.7 & 14.5 & 32.8 & 19.5 & 32.9 & 9.8 & 16.4 & 13.1 & 25.8 & 10.6 & 29.7 \\
    mem0 & Gemma3-12B & 30.8 & \cellcolor{rankfirstblue}\textbf{59.9} & 19.9 & 44.3 & 17.3 & 39.7 & 10.5 & \cellcolor{rankfirstblue}\textbf{27.7} & 12.0 & 26.2 & 49.6 & 80.4 \\
    mem0 & Gemma3-27B & 31.3 & \cellcolor{rankthirdpeach}56.6 & 19.5 & \cellcolor{rankthirdpeach}48.5 & 16.5 & \cellcolor{rankfirstblue}\textbf{45.8} & 5.2 & \cellcolor{rankthirdpeach}18.9 & 7.5 & 21.4 & \cellcolor{rankthirdpeach}87.7 & \cellcolor{ranksecondgreen}\underline{87.5} \\
    \midrule
    mem0* & Gemma3-4B & 15.7 & 30.6 & 9.7 & 15.6 & 17.3 & 27.4 & 8.9 & 10.1 & 11.0 & 21.4 & 11.7 & 35.9 \\
    mem0* & Gemma3-12B & 26.3 & 48.2 & 15.9 & 30.4 & 15.5 & 38.8 & 9.6 & \cellcolor{ranksecondgreen}\underline{19.8} & 10.8 & \cellcolor{ranksecondgreen}\underline{28.4} & 50.2 & \cellcolor{rankthirdpeach}83.0 \\
    mem0* & Gemma3-27B & 30.6 & \cellcolor{ranksecondgreen}\underline{57.1} & 19.2 & 46.9 & 17.4 & \cellcolor{ranksecondgreen}\underline{41.7} & 5.0 & 16.8 & 7.8 & 22.3 & \cellcolor{rankfirstblue}\textbf{91.4} & \cellcolor{rankfirstblue}\textbf{89.9} \\
    \midrule
    mem$\alpha$ & Gemma3-4B & 11.8 & 16.1 & 8.5 & 9.0 & 15.0 & 17.5 & 8.8 & 3.8 & 9.0 & 12.2 & 20.9 & 28.6 \\
    mem$\alpha$ & Gemma3-12B & 22.1 & 41.2 & 10.9 & 18.3 & 15.0 & 32.9 & 9.5 & 16.4 & 8.8 & 21.8 & 62.6 & 81.5 \\
    mem$\alpha$ & Gemma3-27B & 22.7 & 43.0 & 13.3 & 33.0 & 16.6 & \cellcolor{rankthirdpeach}40.4 & 5.2 & 17.7 & 7.6 & 18.8 & \cellcolor{ranksecondgreen}\underline{89.0} & \cellcolor{rankfirstblue}\textbf{89.9} \\
    \midrule
    BM25 & Gemma3-4B & 15.2 & 33.0 & 14.4 & 37.9 & 14.7 & 26.5 & 9.6 & 8.4 & 5.5 & 12.7 & 0.2 & 1.3 \\
    BM25 & Gemma3-12B & 29.1 & 51.8 & 22.7 & \cellcolor{ranksecondgreen}\underline{54.6} & 18.1 & 33.3 & 10.3 & 14.3 & 13.4 & \cellcolor{rankthirdpeach}28.0 & 8.6 & 66.3 \\
    BM25 & Gemma3-27B & 29.5 & 54.2 & 26.1 & \cellcolor{rankfirstblue}\textbf{56.0} & 17.9 & 35.4 & 10.2 & 14.7 & 12.8 & 25.8 & 33.5 & 70.9 \\
    \midrule
    Embedding & Gemma3-4B & 14.1 & 28.6 & 12.5 & 30.6 & 14.8 & 24.5 & 10.1 & 9.2 & 5.6 & 13.1 & 0.4 & 1.8 \\
    Embedding & Gemma3-12B & 25.9 & 48.0 & 20.2 & 46.7 & 17.3 & 31.3 & 10.2 & 16.8 & 12.8 & \cellcolor{rankfirstblue}\textbf{29.3} & 8.6 & 68.9 \\
    Embedding & Gemma3-27B & 26.6 & 48.9 & 22.5 & 44.5 & 18.8 & 34.9 & 10.5 & 16.8 & 12.7 & 27.1 & 39.4 & 72.5 \\
    \bottomrule
  \end{tabular}%
  }
\end{table*}

\clearpage
\onecolumn
\section{Generation Prompts and Examples}
\label{app:generation_prompts}

\begin{promptbox}{Admission Conversation and Summary Prompt}
System:
You are generating a synthetic but clinically grounded conversation for exactly one hospital admission.

Rules:
1. Use only facts supported by the provided admission context.
2. The discharge notes are the backbone of the hospital-course narrative. Use them first.
3. The ordered problem list should shape the main clinical focus.
4. Radiology, procedures, microbiology, and the previous-admission summary are supporting context, not a checklist.
5. Not every packet item needs to appear in the conversation.
6. Speakers must be exactly: Doctor and Patient.
7. Every conversation line must contain:
   - turn number
   - exact timestamp
   - speaker
   - text
8. Timestamps must be monotonically increasing and stay within the admission start and admission end times.
9. Keep the conversation medically plausible and understandable to a patient.
10. Use short natural exchanges, patient questions, clarifications, explanations, reassurance, and follow-up discussion.
11. Do not invent unsupported diagnoses, procedures, test findings, or outcomes.
12. Do not mention the dataset, packet, note types, or that the conversation is synthetic.
13. Generate the conversation first, then write the summary strictly from the generated conversation lines.
14. Every claim in the summary paragraph and every listed problem must be explicitly supported by the conversation lines.
15. Return output that exactly matches the required JSON schema.

User:
Generate one rich doctor-patient conversation for this admission and one admission summary.

Generation guidance:
- Approximate stay length: {stay_days} day(s).
- Recommended turn range: {recommended_turn_range} turns.
- Exact admission window: {admission_start} to {admission_end}.

Return JSON with this shape:
{
  "conversation_lines": [
    {
      "turn_number": 1,
      "time": "YYYY-MM-DD HH:MM:SS",
      "speaker": "Doctor",
      "text": "..."
    }
  ],
  "summary": {
    "admission_start": "YYYY-MM-DD HH:MM:SS",
    "admission_end": "YYYY-MM-DD HH:MM:SS",
    "summary_paragraph": "...",
    "problems": ["...", "..."]
  }
}
\end{promptbox}

\clearpage

\begin{promptbox}{Single-Admission Answerable QA Prompt}
System:
You are generating benchmark-quality hard short-answer question-answer pairs from one hospital-admission conversation between a doctor and a patient.

Rules:
1. Use the provided admission conversation as the evidence source.
2. Generate only answerable questions.
3. Outside medical knowledge is allowed only if it is common, stable, and clinically basic.
4. Use outside knowledge to interpret the context, not to invent unsupported facts.
5. Every question must be hard and should usually require synthesis across multiple turns.
6. Questions must sound natural and be written for benchmark evaluators, not for the patient.
7. Do not use second-person wording like "you" or "your" in the question text.
8. Use third-person phrasing such as "the patient", "the patient's symptoms", or "the doctor" when needed.
9. Every answer must be a short open answer, not yes/no, not multiple choice, and not more than 10 words.
10. Evidence must cite the admission id plus supporting turn_ids.
11. Output valid JSON only.

User:
Generate exactly {question_count} hard answerable short-answer question-answer pairs for this single admission.

Allowed question_type values:
- medical_reasoning
- care_plan_rationale

Return JSON with this exact shape:
{
  "qas": [
    {
      "qa_id": "...",
      "scope": "single_admission",
      "question_type": "medical_reasoning",
      "question": "During the hospitalization from YYYY-MM-DD to YYYY-MM-DD, which condition most likely drove the antibiotic escalation?",
      "answer": "worsening pneumonia",
      "evidence": {
        "admissions": ["..."],
        "turn_ids": [1, 5, 9]
      }
    }
  ]
}
\end{promptbox}

\end{document}